\documentclass[a4paper, 10pt, conference]{ieeeconf}      % Use this line for a4 paper

\IEEEoverridecommandlockouts                              % This command is only needed if 
\usepackage{graphics} % for pdf, bitmapped graphics files
\usepackage{epsfig} % for postscript graphics files
\usepackage{mathptmx} % assumes new font selection scheme installed
\usepackage{times} % assumes new font selection scheme installed
\usepackage{siunitx}
\usepackage{hyperref}
\usepackage{amsmath} % assumes amsmath package installed
\makeatletter
\let\blx@rerun@biber\relax
\makeatother
\DeclareUnicodeCharacter{0301}{}
\usepackage[%
  bibstyle=ieee,
  citestyle=numeric,
  isbn=true,
  doi=false,
  sorting=none,
  url=true,
  defernumbers=true,
  bibencoding=utf8,
  backend=biber,
  maxnames=4
]{biblatex}
\makeatletter
\let\blx@rerun@biber\relax
\makeatother
\makeatletter
\let\blx@rerun@biber\relax
\makeatother
\title{\LARGE \bf
A Study on Simultaneous Use of a Robotic Walker and a Pneumatic Walking Assist Device Designed for PD Patients
}

\author{A. Ali$^{1}$, R. Kawamoto
$^{2}$ and T. Shibata$^{3}$% <-this % stops a space
\thanks{$^{1}$A. Ali is with the Kyushu Institute of Technology,
        Kitakyushu, Fukuoka, Japan
        {\tt\small;
        abdulaleey@gmail.com}}%
\thanks{$^{2}$R. Kawamoto is with the Kyushu Institute of Technology, Kitakyushu, Fukuoka, Japan
        {\tt\small; kawamoto.rikuo867@mail.kyutech.jp}}%
\thanks{$^{3}$T. Shibata is with the Kyushu Institute of Technology,
        Kitakyushu, Fukuoka, Japan
        {\tt\small;
        tom@brain.kyutech.ac.jp}}%
}

\begin{document}

\maketitle
\thispagestyle{empty}
\pagestyle{empty}

%%%%%%%%%%%%%%%%%%%%%%%%%%%%%%%%%%%%%%%%%%%%%%%%%%%%%%%%%%%%%%%%%%%%%%%%%%%%%%%%
\begin{abstract}

Parkinson’s disease (PD) is a common neurodegenerative disease that affects motor and non-motor symptoms. Postural instability and freezing of gait (FOG) are considered motor symptoms of PD resulting in falling. In this study, we investigated the effect of simultaneous use of a robotic walker and a pneumatic walking assist device (PWAD) for PD patients on gait features. The pneumatic actuated artificial muscle on the leg and actuators on the walker produce mutual induced stimulation, allowing the user to suppress FOG and maintain a stable gait pattern while walking. The performance of the proposed system was evaluated by conducting an 8 [m] straight-line walking task by a healthy subject with (a) RW (robotic walker), (b) simultaneous use of an RW and a PWAD, and some gait features for each condition were analyzed. The increasing stride length and decreasing stance phase duration in the gait cycle suggest that simultaneous use of a robotic walker and a pneumatic walking assist device would effectively decrease FOG and maintain a stable gait pattern for PD patients.

\end{abstract}\vspace{2mm}

%%%%%%%%%%%%%%%%%%%%%%%%%%%%%%%%%%%%%%%%%%%%%%%%%%%%%%%%%%%%%%%%%%%%%%%%%%%%%%%%
\section{INTRODUCTION}

Parkinson’s disease (PD) is a progressive neurodegenerative disease involving problems of movement,
emotions, and cognition, affecting more than 10 million people worldwide
\cite{zigmond2014exercise}, and more than 250,000 people are expected to increase in Japan in the future due to the aging population.
\cite{dorsey2018global}.  
%
%% PD is characterized by motor symptoms such as tremors, gait disorders, bradykinesia, and postural
%% instability, as well as non-motor symptoms including autonomic function disorder, sleep disturbance,
%% cognitive and psychiatric disturbances \cite{sveinbjornsdottir2016clinical}. 
%
PD is age-related and increasing due to longer life expectancy. %The number of PD patients in Japan
%is
%
The motor symptoms have an adverse effect on activities of daily life (ADL) which results in
compromising the quality of life (QOL).
Gait disorders of PD include postural instability, festination, freezing of gait (FOG), and
shuffling steps \cite{chen2013gait}. 
Amboni et al. \cite{amboni2015prevalence} suggested that more than half of PD patients experience
FOG; this indicates that FOG is a common symptom of PD patients.\vspace{2mm}

%% Robotic technologies have an essential role in rehabilitation and assistive devices. The most
%% critical rehabilitative application is assisting older people in mobility, as mobility is considered
%% crucial for ADL. The pandemic of coronavirus disease 2019 (COVID-19) forced older people to focus on
%% the infection and lower the priority of care and rehabilitation exercises \cite{kitani2021risk}. The
%% increased mental burden and state of the emergency due to the COVID-19 pandemic has adverse effects
%% on motor, and non-motor symptoms in PD patients \cite{helmich2020impact}. 
%

%% This motivates us to investigate the effect of simultaneous use of a robotic walker and a pneumatic
%% walking assist device to improve FOG, postural instability, and providing rehabilitation to PD
%% patients at home. 

Uchitomi et al. \cite{uchitomi2012interpersonal} developed a virtual robot Walk-Mate based on
external auditory stimulation to stabilize gait in PD patients. External cueing, such as audio or
visual stimulation, is considered a clinically accepted strategy to overcome FOG
\cite{rocha2014effects}, but the effects of cueing have been reported to diminish with time
\cite{ginis2018cueing}.
Some patients with poor rhythmic abilities do not respond to the cues or gait may also worsen with
cueing \cite{dalla2018individualization}. GONDOLA Automated Mechanical Peripheral Stimulation (AMPS)
\cite{gondola_2021} was developed for the suppression of FOG to improve gait speed but cannot
provide stability in a standing posture \cite{pinto2018automated}. The study \cite{wheelchair} reported that using a pedal-powered wheelchair COGY, PD patients can
demonstrate regular and fast pedal movements without having FOG.
Motivated by the
results with COGY, Higuchi, et al. previously developed a novel pneumatic walking assist device that
gives self-paced and self-induced tactile and force feedback on the knee, and showed its efficacy on
the double support phase during the gait of PD patients \cite{higuchi2020effects}\vspace{2mm}.

%Our research focuses on proposing a system for PD patients whose condition does not require a
%wheelchair. 

In this study, we further propose an integrated system consisting of a robotic walker and a pneumatic
walking assist device for improving the gait by mutual induced stimulation, and our preliminary
study with a healthy subject shows the plausibility of the proposed system on the gait features.\vspace{2mm} 

%
%The proposed system helps in decreasing FOG and provides stable walking to PD patients. 

%% The use of a walker in rehabilitation and assistance can increase confidence by
%% providing a smooth gait and is also considered a good choice for PD patients having postural
%% instability \cite{kegelmeyer2013assistive}.\vspace{2mm}

The paper is organized as follows. Section 2 describes the detailed description of the proposed
system. In section 3, the information about the experiment is presented. Section 4 shows
experimental results along with the discussion. Conclusion and future works are presented in section
5.\vspace{2mm}

\section{PROPOSED SYSTEM}

\subsection{Robotic Walker}

\begin{figure}[ht!]
 \centering
  \includegraphics[width=0.44\textwidth]{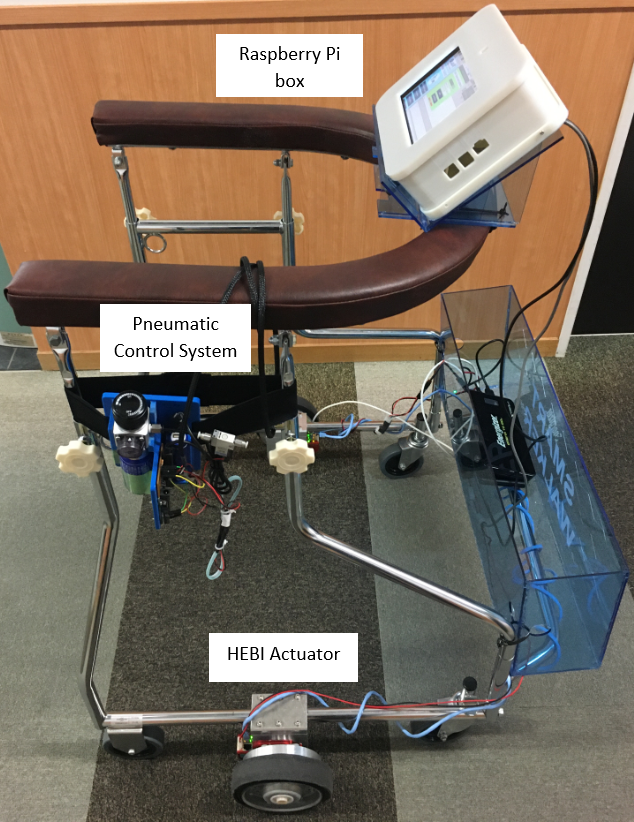}
  % \vspace*{-4mm}
  \caption{Robotic walker with pneumatic walking assist device}
  \label{fig: fig0}
  \vspace{-3mm}
\end{figure}

The developed system is an improved form of UPS-PD \cite{higuchi2019gait} with the robotic walker. To make the prototype, a simple rollator walker was modified. Two additional wheels of diameter 15 [cm] were added to the middle of the walker structure base as shown in Fig. \ref{fig: fig0}. The wheels were attached to HEBI X5-1 actuators having continuous torque of 1.2 [Nm]. The actuators can communicate using ethernet and provide induced stimulation when activated. The actuators can sense commanded and feedback parameters like position, velocity and, torque. Raspberry Pi box consists of raspberry pi 3B+, touch screen, mic, and mini-speaker. Two feedback controllers were implemented to control the motion and braking system of the robotic walker. The system can be operated manually from the user interface or through voice command. The user needs to set speed and distance before operating the robotic walker. The user can apply the braking system during the session by a voice command of "STOP" or emergency button. The robotic walker generates induced stimulation, which helps in decreasing FOG and providing a stable gait pattern.\vspace{2mm}

\begin{figure}[ht!]
 \centering
  \includegraphics[width=0.44\textwidth]{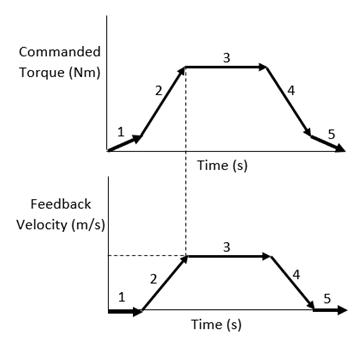}
   \vspace*{-2mm}
  \caption{Torque-Velocity regions during the motion of the robotic walker}
  \label{fig: fig1}
  \vspace{-3mm}
\end{figure}

The velocity feedback controller was implemented to control the motion of the robotic walker. Eq. \eqref{eqn: eq1} shows the control signal, $\tau_t$ (commanded torque at time $t$) of the velocity feedback controller that commands the walker's motion. During the motion of the walker, torque behaves like a control parameter against $v_{ref}$ (referenced feedback velocity) by measuring ${c}_{t}$ (feedback velocity at time $t$). The controller gains were empirically determined. The velocity feedback controller can be analyzed by dividing the robotic walker's motion into five regions based on velocity characteristics, as shown in Fig. \ref{fig: fig1}.\vspace{2mm}

The first region is known as the positive neutral region. To get out of the first region, the user needs to push the robotic walker that generates torque in the wheels. The second region is the accelerating region in which the motion of the robotic walker starts. The torque is increased gradually to reach the desired velocity, and after that robotic walker enters into a third region known as the constant velocity region. A dotted horizontal line in Fig. \ref{fig: fig1} represents the desired velocity. The fourth region, the deaccelerating region, is triggered when the set distance is about to complete. The last region termed the negative neutral region moves actuators to complete rest.\vspace{2mm} 

\vspace{2mm}

\begin{equation}
\tau_t = \tau_{t-1} + 
\begin{cases}
   +0.03 & \text{if} \ (v_{ref} - {c}_t) > 0.05\\
   -0.03 & \text{if} \ (c_{t} - v_{ref}) > 0.05\\
   \ 0, & \text{otherwise} \\
\end{cases}
\label{eqn: eq1}
\end{equation}

\begin{equation}
p_t = p_{t-1} + (p_{ref} - b_{t-1})
\label{eqn: eq2}
\end{equation}
\vspace{-1mm}

The position impedance controller was implemented, which acts as a braking system for the robotic walker. In this controller, position acts as a control parameter and results in the locking of the wheels. Eq. \eqref{eqn: eq2} shows the control signal, $p_t$ (commanded position at time $t$) for the position-based impedance controller. $p_{ref}$ is the reference position recorded when the stop command is activated. $b_{t-1}$ is the position feedback signal provided by actuators at time $t-1$.\vspace{2mm}

\subsection{Pneumatic Walking Assist Device}
\begin{figure}[ht!]
 \centering
  \includegraphics[width=0.40\textwidth]{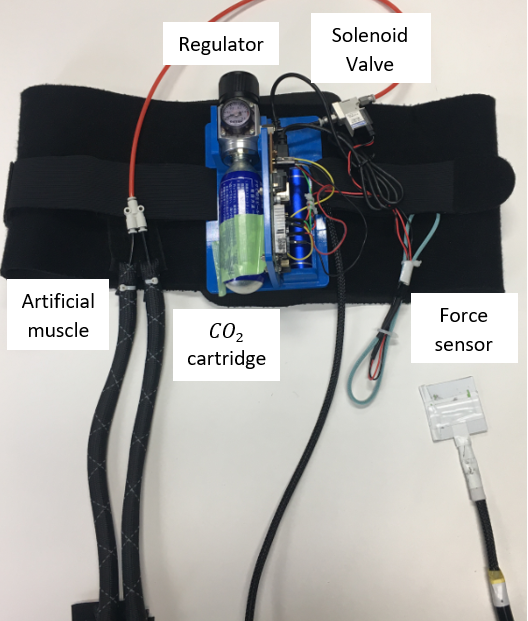}
   \vspace*{-2mm}
  \caption{Pneumatic walking assist device}
  \label{fig: fig2}
  \vspace{-3mm}
\end{figure}

The pneumatic walking assist device has two parts; a pneumatic control system and an artificial muscle. Fig. \ref{fig: fig2} shows a pneumatic walking assist device for PD patients. The pneumatic control system consisted of a $CO_2$ cartridge, regulator, solenoid valve, and Arduino were attached to the robotic walker. Arduino and solenoid valve were used to control gas influx from the $CO_2$ gas cartridge into two parallelly connected artificial muscles.\vspace{1mm}

\vspace{3mm}

\begin{figure}[ht!]
 \centering
  \includegraphics[width=0.44\textwidth]{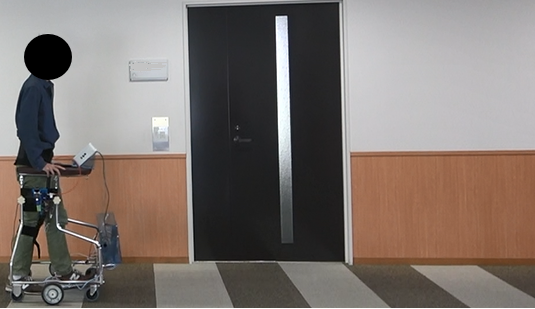}
   \vspace*{-2mm}
  \caption{Simultaneous use of the robot walker and the pneumatic walking assist device in the experiment}
  \label{fig: fig3}
  \vspace{-3mm}
\end{figure}\vspace{3mm}

The artificial muscles were attached to the right leg of the subject. The force sensor (FSR-406) was used on the left shoe's sole to detect left foot ground contact. The artificial muscles were activated when left-foot ground contact was detected to produce induced stimulation, thereby providing walking assistance. The maximum $CO_2$ gas pressure was adjusted to 0.3 [MPa] by the regulator during the operation. \vspace{-2mm}

\section{Experiment}
The PD patients who experience FOG show more stance phase percentage in stride than healthy people
\cite{pozzi2019freezing}. 
The stance phase is the period during which the foot is in contact with ground support.
The FOG in PD patients affects stride length and gait velocity \cite{chee2009gait}.
The PD patients with increased stance phase percentage and decreased stride length are considered at
high fall risk \cite{kwon2018comparison}.
The simultaneous use of a robotic walker and a pneumatic assistive device generates mutual induced
stimulation to decrease the FOG episodes in PD patients.\vspace{2mm} 

The experiment was conducted with a 23-year-old healthy subject. The subject was asked to walk at 8 [m] straight path under two different conditions; condition {A} and condition {B}. In condition {A} the walking task is performed using a robotic walker. In condition {B} the walking task is performed with simultaneous use of a robotic walker and a pneumatic walking assist device. The experiment at each condition was performed twice. The velocity was set at 0.5 [m/s] for all trials. The experiment investigated the effect of simultaneous use of a robotic walker and pneumatic assistive device on different gait features. The PEDAR \cite{novel.de_2021} was used to analyze gait features. Fig. \ref{fig: fig3} shows experimental view. The gait time and step count in both conditions were analyzed. The stance phase percentage in stride and swing phase percentage in stride for both conditions were compared using t-test analysis. To ensure obtaining a steady gait for analysis, the first two steps were excluded from the analysis. The stance phase percentage and swing phase percentage in stride were recorded separately for the left and right leg.\vspace{1mm}

\begin{figure}[ht!]
 \centering
  \includegraphics[width=0.45\textwidth]{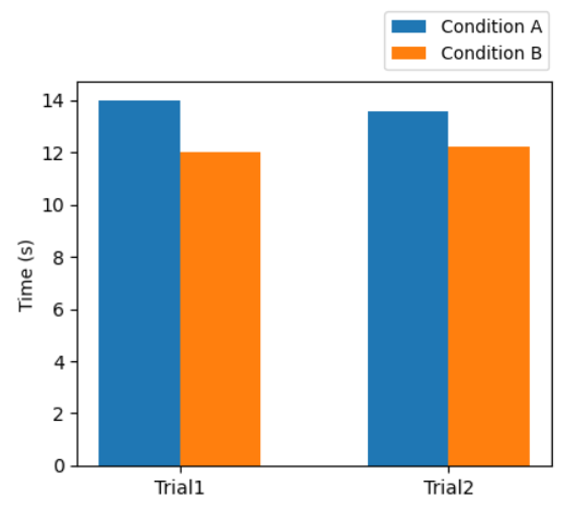}
   \vspace*{-2mm}
  \caption{Gait duration for 8 [m] walking task}
  \label{fig: fig4}
  \vspace{-3mm}
\end{figure}\vspace{2mm}

The Shapiro-Wilk test was used to check the normality of data, and then the t-test was performed for both the average stance phase percentage and average swing phase percentage to investigate the difference between condition A and condition B. The chosen significance level was 5 [\%] in the t-test. During the 8 [m] walking task, step count, gait duration, stance phase percentage in stride, and swing phase percentage in stride were recorded.\vspace{1mm}

\section{Results}
\begin{figure}[ht!]
 \centering
  \includegraphics[width=0.45\textwidth]{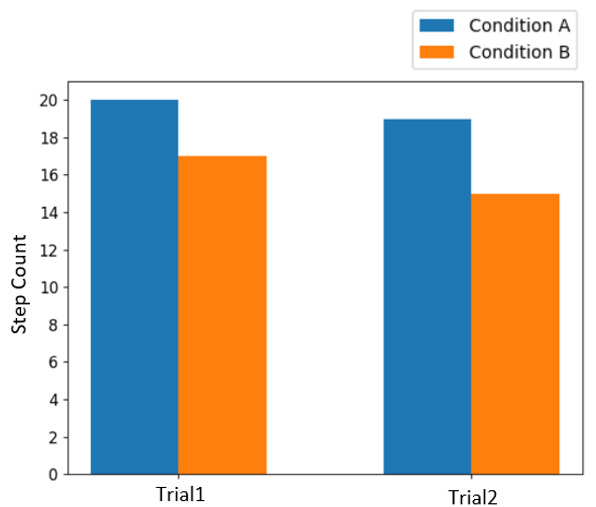}
   \vspace*{-2mm}
  \caption{Number of Steps}
  \label{fig: fig5}
  \vspace{-3mm}
\end{figure}

\begin{figure}[ht!]
 \centering
  \includegraphics[width=0.45\textwidth]{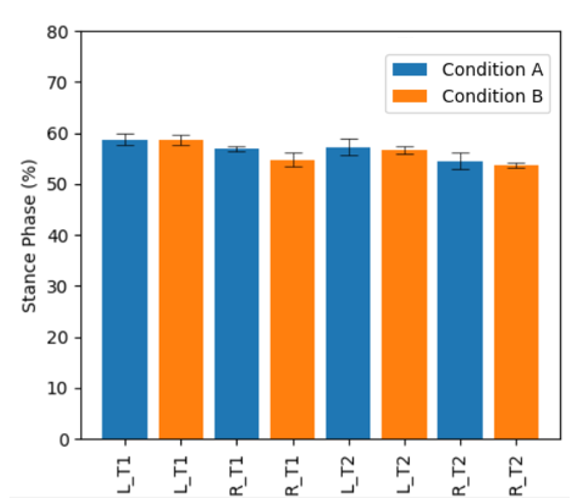}
   \vspace*{-2mm}
  \caption{Average stance phase percentage}
  \label{fig: fig6}
  \vspace{-3mm}
\end{figure}

\begin{figure}[ht!]
 \centering
  \includegraphics[width=0.45\textwidth]{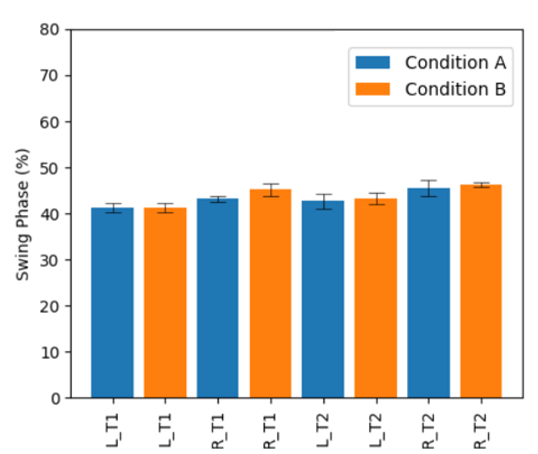}
  \vspace*{-2mm}
  \caption{Average swing phase percentage}
  \label{fig: fig7}
  \vspace{-3mm}
\end{figure}

Fig. \ref{fig: fig4} shows gait duration for the 8 [m] walking task in both conditions. When the subject performed the walking task with condition {B}, a decrease in gait duration was observed in both trials. In Fig. \ref{fig: fig6} and Fig. \ref{fig: fig7}, T1 and T2 represent trial1 and trial2 respectively. L and R mean left and right leg, respectively. Fig. \ref{fig: fig5} shows that step counts were also decreased in condition {B}. The decrease in the number of steps and gait speed indicates that the simultaneous use of both devices increases the stride length. Fig. \ref{fig: fig6} shows the average stance phase percentage in the gait cycle for each leg.\vspace{2mm}

\vspace{3mm}
\vspace{2mm}

\vspace{2mm}

 The stance phase percentage in T1 for the left leg remained 58 [\%] in both conditions. The stance phase percentage for the right leg in T1 was decreased by 2 [\%] in condition {B}. In T2, the stance phase percentage for the right leg could be decreased by 1 [\%] in condition {B}. Fig. \ref{fig: fig7} could show an increase in swing phase percentage for the right leg in T1. The stance phase and swing phase percentage could remain the same for the left leg in both conditions because the artificial muscle was only attached to the subject's right leg. According to the t-test, there was no significant difference between both conditions. A possible reason would be that the experiment was conducted with a healthy subject who did not experience FOG. The potential decrease in step count and gait duration could suggest that stride length was increased in condition {B}. The potential increase in stride length and decreasing stance phase percentage could suggest that simultaneous use of a robotic walker and pneumatic walking assist device can decrease FOG in PD patients.\vspace{0.5mm}

\vspace{2mm}

\section{CONCLUSION}
In the current study, we presented the integrated system consisting of a robotic walker and pneumatic walking assist device and investigated the effect of both devices' simultaneous use on the gait features of the healthy subject. An improvement in gait speed was observed due to mutual induced stimulation. The experimental results indicate that both devices' simultaneous use has no worsening effects on the gait features. The potential increase in the stride length and decrease in the stance phase percentage in the gait cycle could suggest that simultaneous use of both devices can decrease FOG in PD patients.\vspace{2mm}

In the future, a pneumatic control system with a multi-cartridge will be implemented to increase the use time of the pneumatic walking assist device. The robotic walker's braking system will also be modified by generating a feedback signal from the force sensor used on the left sole. The improved braking system will be more effective in reacting to emergencies and decrease fall risks. Since the subject used in this study was healthy, the result obtained from this study served as a first step in developing the integrated system consisting of a robotic walker and pneumatic walking assist device, setting evaluation protocols, and determining the effect of the simultaneous use of both devices on different gait features. Future work will focus on evaluating the integrated system with PD patients.\vspace{2mm}


%%%%%%%%%%%%%%%%%%%%%%%%%%%%%%%%%%%%%%%%%%%%%%%%%%%%%%%%%%%%%%%%%%%%%%%%%%%%%%%%

%%%%%%%%%%%%%%%%%%%%%%%%%%%%%%%%%%%%%%%%%%%%%%%%%%%%%%%%%%%%%%%%%%%%%%%%%%%%%%%%

%%%%%%%%%%%%%%%%%%%%%%%%%%%%%%%%%%%%%%%%%%%%%%%%%%%%%%%%%%%%%%%%%%%%%%%%%%%%%%%%

\setlength\bibitemsep{0pt}   % length between two different entries
\renewcommand*{\bibfont}{\small} 
\printbibliography

\footnotesize

\normalsize
\end{document}